\documentclass[10pt,twocolumn,letterpaper]{article}

\usepackage{cvpr}
\usepackage{times}
\usepackage{epsfig}
\usepackage{graphicx}
\usepackage{amsmath}
\usepackage{amssymb}
\usepackage{bbm}
\usepackage{array}
\usepackage{multirow}
\usepackage{caption} 
\usepackage{enumitem}

\usepackage[pagebackref=true,breaklinks=true,letterpaper=true,colorlinks,bookmarks=false]{hyperref}
\cvprfinalcopy

\def\cvprPaperID{1}

\ifcvprfinal\fi
\begin{document}

\title{Complexer-YOLO: Real-Time 3D Object Detection\\and Tracking on Semantic Point Clouds}

\author{Martin Simon, Karl Amende, Andrea Kraus, Jens Honer,\\
Timo S\"amann, Hauke Kaulbersch and Stefan Milz\\
Valeo Schalter und Sensoren GmbH\\
{\tt\small \{firstname.lastname\}@valeo.com}
\and
Horst Michael Gross\\
Ilmenau University of Technology\\
{\tt\small horst-michael.gross@tu-ilmenau.de}
}

\maketitle

\begin{abstract}
Accurate detection of 3D objects is a fundamental problem in computer vision and has an enormous impact on autonomous cars, augmented/virtual reality and many applications in robotics. In this work we present a novel fusion of neural network based state-of-the-art 3D detector and visual semantic segmentation in the context of autonomous driving. Additionally, we introduce Scale-Rotation-Translation score (SRTs), a fast and highly parameterizable evaluation metric for comparison of object detections, which speeds up our inference time up to 20\% and halves training time. On top, we apply state-of-the-art online multi target feature tracking on the object measurements to further increase accuracy and robustness utilizing temporal information. Our experiments on KITTI show that we achieve same results as state-of-the-art in all related categories, while maintaining the performance and accuracy trade-off and still run in real-time. Furthermore, our model is the first one that fuses visual semantic with 3D object detection.
\end{abstract}

\begin{figure*}
\centering
\includegraphics[width=\textwidth]{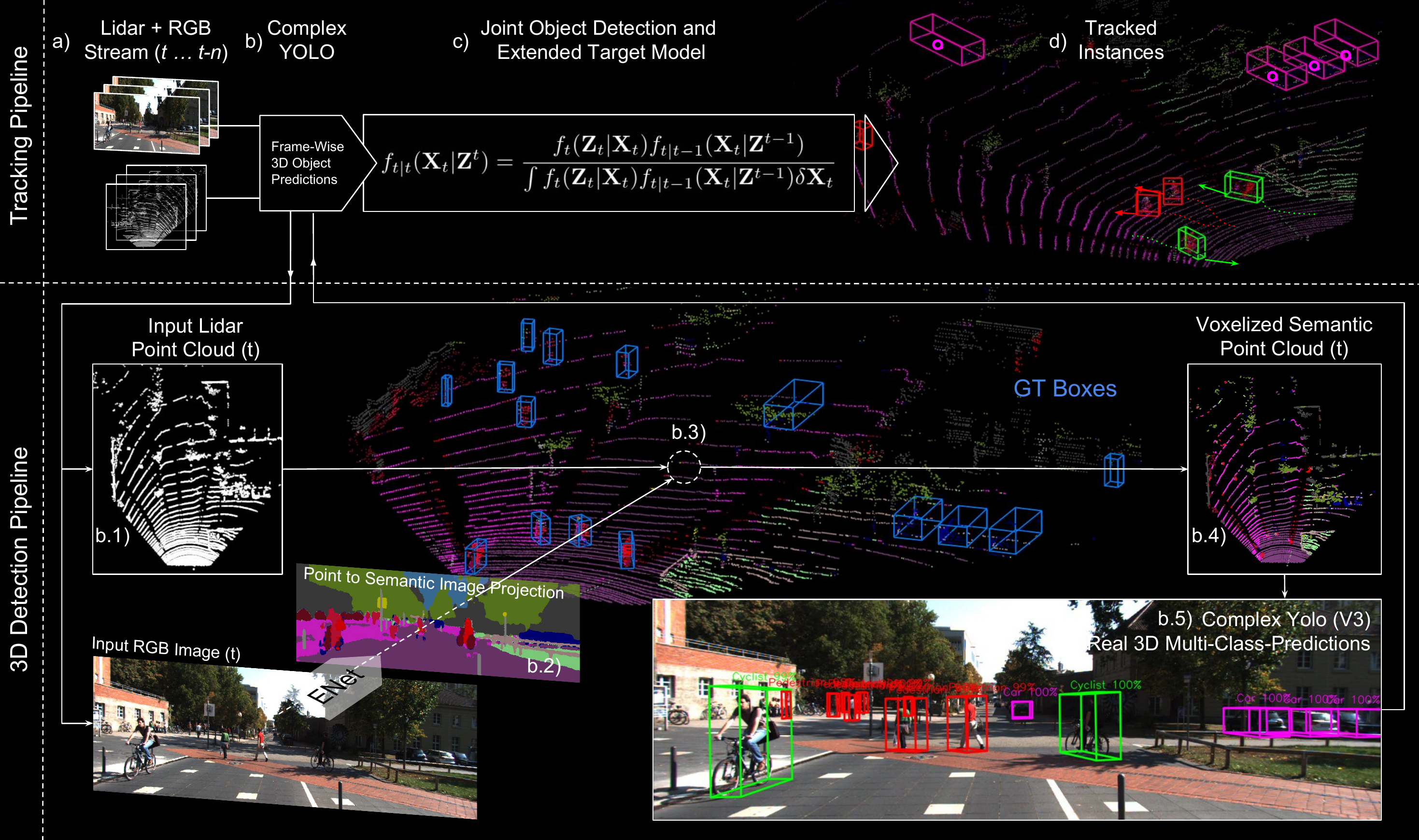}
\caption{The Complexer-YOLO processing pipeline: We present a novel and complete 3D Detection (b.1-5) and Tracking pipeline (a,b,c,d,e) on Point Clouds in Real-Time. The Tracking-Pipeline is composed by: (a) Lidar + RGB frame grabbing from stream, (b) Frame-wise Complex-YOLO 3D Multiclass predictions, (c) Joint Object and extended Target Model for feature Tracking and (d) 3D object instance tracking within the environmental model. In detail (b) is composed by: (1) The Voxelization of the Lidar frame, (2) the Semantic Segmentation of the RGB image with the aid of ENet, (3) the Point-wise classification by Lidar to Semantic-Image backprojection, (4) the generation of the Semantic Voxel Grid and finally (5): The real 3D Complex YOLO for 3D Multi-class predictions. (see Fig.~\ref{fig:Design} for more details)  }
\label{fig:GAN}
\end{figure*}

\section{Introduction}
Over the last few years self-driving cars got more and more into the focus of the automotive industry as well as new mobility players. Today, commercial vehicles already offer manifold automation like assisted or automated parking, adaptive cruise control and even highway pilots. To reach the full level of automation, they require a very precise system of environmental perception, working for every conceivable scenario. Additionally, real world scenarios strictly require real-time performance.

Recent vehicles are equipped with multiple different kind of sensors like ultrasonics, radar, cameras and Lidar (light detection and ranging) as well. With the help of redundancy and sensor fusion, relevant reliability and safety can be achieved. These circumstances significantly boosted the rapid development of sensor technology and the growth of artificial intelligence algorithms for fundamental tasks like object detection and semantic segmentation. 

\begin{figure*}[!htb]
\centering
\includegraphics[width=0.80\textwidth]{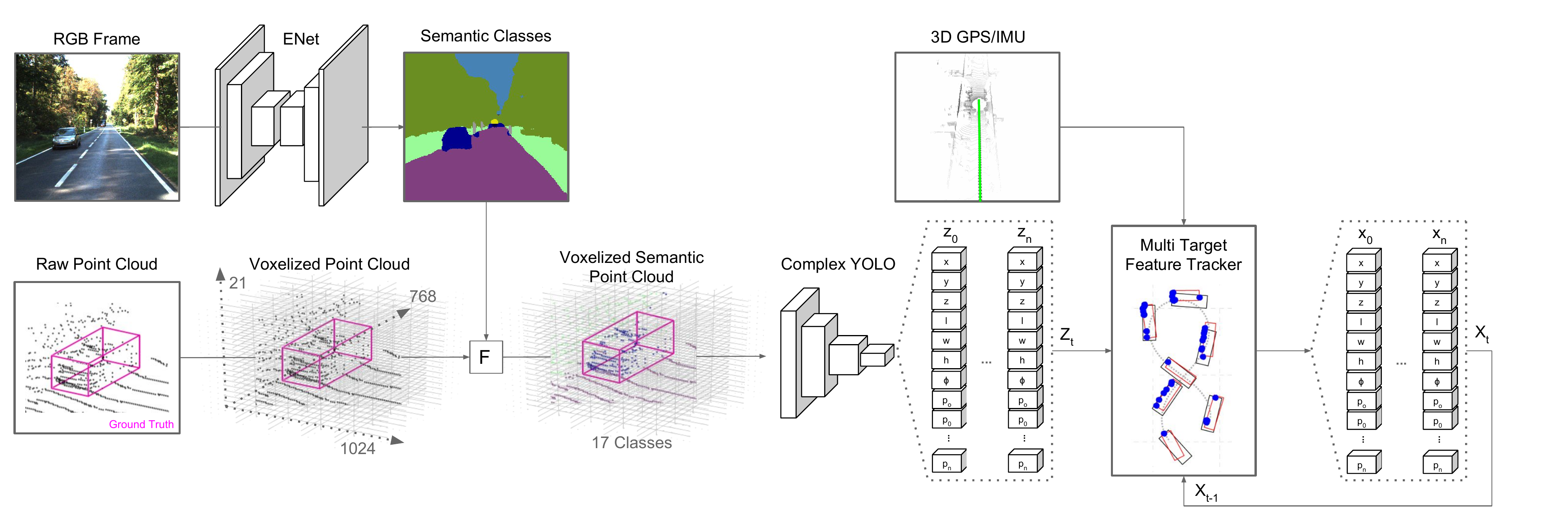}
\caption{Overview of our architecture.}
\label{fig:Design}
\end{figure*}

Many modern approaches for these tasks use camera, Lidar or combine both. Compared to camera images, there are some difficulties dealing with Lidar point cloud data. Such point clouds are unordered, sparse and have a highly varying density due to the non-uniform sampling of the 3D space, occlusion and reflection. On the other hand, they offer way higher spatial accuracy and reliable depth information. Therefore, Lidar is more common in the context of autonomous driving. In this paper, we propose Complexer-YOLO, a real-time 3D object detection and tracking on semantic point clouds (see Fig. \ref{fig:GAN}, \ref{fig:Design}). The main contributions are:
\begin{itemize}
\item \textbf{Visual Class Features}: Incorporation of visual point-wise Class-Features generated by fast camera-based Semantic Segmentation~\cite{EfficientSemantic}.
\item \textbf{Voxelized Input}: Extension of Complex-YOLO \cite{ComplexYolo} processing voxelized input features with a variable  depth of dimension instead of fixed RGB-maps.
\item \textbf{Real 3D prediction}: Extension of the regression network to predict 3D box heights and z-offsets to treat targets in three dimensions.
\item \textbf{Scale-Rotation-Translation score (SRTs)}: We introduce \textit{SRTs}, a new validation metric for 3D boxes, notably faster than intersection over union (IoU), considering the 3DoF pose of the detected object including the yaw angle such as width, height and length. 
\item \textbf{Multitarget-Tracking}: Application of an Online feature tracker decoupled from the detection network, enabling time depending tracking and target instantiation based on realistic, physical assumptions.
\item \textbf{Realtime capability}: We present a complete novel tracking pipeline with an outstanding overall real-time capability, despite state-of-the-art results on semantic segmentation, 3D object detection such as Multitarget-Tracking. The pipeline can be directly brought into every self-driving cars percepting urban scenes.
\end{itemize}

\section{Related Work}
In this section, we provide an overview of convolutional neural network (CNN) based object detection, semantic segmentation and multi target tracking.

\subsection{2D Object Detection}
Over the last few years many methods for robust and accurate object detection using CNN have been developed. Starting in 2D space on single images, two-stage detectors \cite{FasterRCNN,Mask-RCNN} and one-stage detectors \cite{YOLO,SSD,YOLOv2,RetinaNet,YOLOv3,SqueezeNet} achieved state-of-the-art results, targeting the output of located 2D bounding boxes. Typically, two-stage detectors exploit object proposals and utilize region of interests (RoI) with the help of region proposal networks (RPN) in a first step. Afterwards, they generate the final object predictions using calculated features over the proposed RoIs. As a trade-off for runtime, one-stage detectors skip the proposal generation step and directly output the final object detections. They are usually capable of real-time performance, but mainly outperformed by two-stage detectors in terms of accuracy. YOLOv3 \cite{YOLOv3}, one of the one-stage detectors, combines findings from \cite{YOLO,YOLOv2,ResNet,FeaturePyramidNetworks}. It divides the image into a sparse grid, performs multi-scale feature extraction and directly outputs object predictions per grid cell, using dimension clusters as anchor boxes \cite{YOLOv2}.

\subsection{3D Object Detection}
Although CNNs were originally designed for image processing, they became a key component for 3D object detection as well. First ideas were to use stereo images as input \cite{3DObjectProposalStereo}. Followed by \cite{3DFullyConvVehicleDet} and \cite{Vote3Deep}, where 3D convolutions were applied to a voxelized representation of point cloud data and features extracted using 3D CNNs \cite{3DConv}, respectively. In \cite{VoxelNet}, voxel feature encoding was introduced and again processed by CNN to predict objects. Furthermore, VeloFCN \cite{VeloFCN} created depth maps with the help of front-view projections of 3D point clouds and applied CNN. In contrast, MV3D \cite{MV3D} merged image input with a multi-view representation of point cloud data projected into 2D space. Alternatively, \cite{AVOD} aggregated features from image and birdseye-view representation of point clouds. Another method to fuse camera and Lidar inputs was explored in \cite{FrustumPointNets}. In a first step, sub point clouds were extracted in viewing frustums detected by a 2D CNN. Afterwards, a PointNet \cite{PointNet} predicts 3D objects within the frustum point clouds. Recently, PointNet was also used in \cite{PointFusion} in combination with 2D CNN and a fusion network. Further similar approaches using birdseye-view representations of point clouds were \cite{FastAndFurious,PIXOR,ComplexYolo,BirdNet}. 

\subsection{Semantic Segmentation}
The goal of semantic segmentation is to classify each pixel of an image into a predefined class. This task is typically achieved by CNNs. Several widely used network architectures have been introduced, e.g. \cite{RecurrentNeuralNetworks,ENet,ERFNet,RefineNet}. Similar to the object detection task, there is a trade-off between accuracy and runtime. Therefore, approaches like convolutional factorization, e.g. applied in \cite{MobileNets,ESPNet}, quantization \cite{XNOR}, pruning \cite{ChannelPruning} and dilated convolutions, e.g. applied in \cite{DilatedConvolutions}, came up. ENet \cite{ENet}, one of the most efficient models used a special encoder-decoder structure to highly reduce computational effort. Recently, \cite{EfficientSemantic} applied the Channel Pruning method \cite{ChannelPruning} to the ENet to make it more efficient.

\subsection{Multi Target Tracking} \label{sec_multi_target_tracking}
The task of multi-object tracking (MOT) is usually solved in two phases. First, an algorithm detects objects of interests and second, identical objects in different frames are associated. A widespread approach is using global information about the detections \cite{MultiClassMultiObjectTracking,TrackingByDetection}. In contrast to this, online approaches don't have any knowledge of future frames. With this characteristic they have one significant advantage: they are usable in real-world scenarios.  Recent work focused especially on tracking of 2D objects from camera input \cite{MultiObjectTrackingDeepLearningPMBMFilterung,BeyondPixelsMultiObjectTracking}. Online multi-target 3D object tracking based on detections from algorithms with point cloud inputs aren't popular until now.
The basics of the Labeled Multi-Bernoulli Filter, which we use for multi target tracking, are explained in the following.

The state $x_t^i$ of the $i$th target at discrete time $t$ is a random variable.  
The set of all targets at time step $t$ is a subset of the state space $\mathbb{X}$ and then denoted by
\begin{align} \label{eq:target_set}
\mathbf{X}_t = \left\{ x_t^i  \right\}_{i=1}^{N^x_t} \subset \mathbb{X}.
\end{align}
In turn, the set cardinality $N^x_t = |\mathbf{X}_t|$ at time $t$ is a discrete random variable. 

The set of all measurements at time $t$ is again modeled as a random set with set cardinality $N^z_t = |\mathbf{Z}_t|$ and denoted by
\begin{align} \label{eq_measurement_set}
\mathbf{Z}_t = \left\{ z_t^i  \right\}_{i=1}^{N^z_t} \subset \mathbb{Z}.
\end{align}

Each individual measurement $z_t^i$ is either target-generated or clutter. Yet the true origin is assumed unknown. Further, the set of all measurements up to and including the time step $t$ is denoted by
\begin{align}
\mathbf{Z}^t = \bigcup_{\tau = 1}^{t}  \mathbf{Z}_\tau.
\end{align}
Both above sets are without order, i.e. the particular choice of indices is arbitrary. Targets and measurements are modeled as Labeled Multi-Bernoulli Random Finite Sets (LMB RFS) as proposed in \cite{GLMB}. 
A Bernoulli RFS is a set that is either empty with probability $1-r$ or contains a single element. As in \cite{LabeledMultiBernoulli}, the probability density of a Bernoulli RFS may be written as
\begin{align}
\pi(X) &=  \begin{cases}
1-r, & \text{if } X = \emptyset, \\
r \: p(x), & \text{if } X = \{x\}
\end{cases}\label{eq_prp_desity_RFS}
\end{align}
with $p(\cdot)$ a spatial distribution on $\mathbb{X}$. A Multi-Bernoulli RFS is then the union of independent Bernoulli RFSs, i.e. $X_\mathrm{MB} = \bigcup_i X_\mathrm{B}^{(i)}$. In turn a Multi-Bernoulli RFS is well-defined by the parameters $\{r^{(i)}, p^{(i)}\}_i$.

Labeled RFS allow the estimation of both the targets' state and their individual trajectories. For this reason the target state is extended by a label $l \in \mathbb{L}$, i.e. each single target state is given by $\mathbf{x} = (x, l)$ and in turn the multi-target state $\mathbf{X}$ lives on the product space $\mathbb{X} \times \mathbb{L}$ with $\mathbb{L}$ a discrete space. Note that this definition does not enforce the labels $l$ to be distinct. \cite{RandomFiniteSet} introduced the so called distinct label indicator
\begin{align}
\Delta (\mathbf{X}) &:= \delta_{|\mathbf{X}|} (|\mathcal{L}(\mathbf{X})|)
\end{align}
that enforces the cardinality of $\mathbf{X}$ to be identical to the cardinality of the projection $\mathcal{L}(\mathbf{X}) = \{\mathcal{L}(\mathbf{x}) :\mathbf{x} \in \mathbf{X} \}$, $\mathcal{L}(\mathbf{x}) = l$.
Together with Eq. \ref{eq_prp_desity_RFS}, it follows that the probability density of a LMB RFS is well-defined by the parameter set $\{r^{(l)}, p^{(l)}\}_{l \in \mathbb{L}}$ and the cardinality distribution yields
\begin{align} \label{eq_card_dist}
\rho(n) = \prod_{i \in \mathbb{L}} (1 - r^{(i)}) \sum_{L \in \mathcal{F}_n(\mathbb{L})} \prod_{l \in L} \frac{r^{(l)}}{1-r^{(l)}} 
\end{align}
with $\mathcal{F}_n(\mathbb{L})$ the set of all subsets of $\mathbb{L}$ containing $n$ elements.

The core objective of the multi-target tracking is to approximate the multi-target distribution $f_{t|t}(\mathbf{X}_t|\mathbf{Z}^t)$ in each time step $t$. This is achieved with the multi-target Bayes filter,
\begin{align}
f_{t|t}(\mathbf{X}_t|\mathbf{Z}^t) &= 
\frac{f_t(\mathbf{Z}_t|\mathbf{X}_t) f_{t|t - 1}(\mathbf{X}_t|\mathbf{Z}^{t - 1})}
{\int f_t(\mathbf{Z}_t|\mathbf{X}_t) f_{t|t - 1}(\mathbf{X}_t|\mathbf{Z}^{t - 1}) \delta \mathbf{X}_t}
\end{align}
and the Chapman-Kolmogorov prediction
\begin{align}
f_{t+1|t}(\mathbf{X}_{t+1}|\mathbf{Z}^t) &= 
\int f_{t+1|t}(\mathbf{X}_{t+1}|\mathbf{X}_t) f_{t|t}(\mathbf{X}_t|\mathbf{Z}^{t}) \delta \mathbf{X}_t \label{eq_Chapman-Kolmogorov}
\end{align}
with $f_t(\mathbf{Z}_t|\mathbf{X}_t)$ the multi-target measurement set density and $f_{t+1|t}(\mathbf{X}_{t+1}|\mathbf{X}_t)$ the multi-target transition density.

\section{Joint Detection and Extended Target Model}

\subsection{Point Cloud Preprocessing}
First, we generate a semantic segmentation map of the front camera images using the efficient model from \cite{EfficientSemantic}, pre-trained on \cite{Cityscapes} and fine tuned on KITTI \cite{KITTI}. Second, we quantize the point cloud to a 3D voxel representation, which is able to contain certain features of the points that lie within such a voxel. Our region of interest of the point cloud is set to $[0,60]m \times [-40,40]m \times [-2.73,1.27]m$ in sensor coordinates, according to the KITTI \cite{KITTI} dataset. We chose a resolution of $768 \times 1024 \times 21$ resulting in approximately $0.08m \times 0.08m \times 0.19m$ per cell. Each voxel, where at least one point exists inside its 3D space and is visible to the front camera, is filled with a normalized class value extracted from the semantic map in range $[1, 2]$. Therefore, we project all relevant points into the image using calibrations from \cite{KITTI} and \textit{argmax} over the frequency of all resolved classes. In this way, contextual information with visual features are passed into our voxel map, which is especially helpful for higher ranges with low density of points.

\subsection{Voxel based Complex-YOLO}
We use the full detection pipeline introduced in \cite{ComplexYolo}, but exchange the input map for our voxel representation. Inspired by \cite{YOLOv3}, we exchange max-pooling layers by convolutions with stride $2$ and add residual connections. Altogether, we have $49$ convolutional layers. Additionally, we add object height $h$ and ground offset $z$ as target regression parameters and incorporate both into the multi-part loss function.
\begin{equation}
\begin{split}
\mathcal{L} = &\mathcal{L}_{\text{Euler}}\\
&+ \lambda_{\text{coord}}\sum_{i=0}^{S^2}\sum_{j=0}^B\mathbbm{1}_{ij}^{obj} \left[ (h_i - \hat{h}_i)^2 + (z_i - \hat{z}_i)^2 \right]
\end{split}
\label{equ:loss}
\end{equation}

Usually IoU is used to compare detection and ground truth during the training process. However, it has drawbacks when comparing rotated bounding boxes. If two boxes are compared with the same size and position and an angle difference of $\pi$ the IoU between these two boxes is $1$, which means they match perfectly. This is obviously not the case since the angle between the two boxes has the maximum difference it can have. So while training a network it is not penalized and even encouraged by predicting boxes like these. This leads to wrong predictions for the object orientation. Also calculating an exact IoU for rotated bounding boxes in 3D space is a time consuming task.

To overcome these two problems we introduce a new highly parameterizable simple evaluation metric called Scaling-Rotation-Translation score (\textit{SRTs}). The \textit{SRTs} is based on the fact, that given two arbitrary 3D objects of the same shape, one can be transformed into the other using a transformation. Therefore, we can define a score $S_{srt}$ as composite of independent scores for scaling $S_s$, rotation $S_r$ and translation $S_t$ with

\begin{equation}
S_s = 1 - \min\Biggl(\frac{|1-s_x| + |1-s_y| + |1-s_z|}{w_s}, 1\Biggr)
\end{equation}

\begin{equation}
S_r = \max\Bigl(0, 1 - \frac{\theta}{w_r\pi}\Bigr), \quad w_r \in (0, 1]
\end{equation}

\begin{equation}
r_i = \frac{d_i \cdot w_{t}}{2}, \quad i \in \{1,2\}
\end{equation}
\begin{equation}
S_t = \max\Bigl(0, \frac{r_1 + r_2 - t}{r_1 + r_2}\Bigr)
\end{equation}

\begin{equation}
p_t = \begin{cases} 0,& \text{if } r_1 + r_2 < t\\
1,& \text{otherwise}
\end{cases}
\end{equation}

where $s_{x,y,z}$ denotes size ratios in $x$, $y$, $z$ directions, $\theta$ denotes the difference of the yaw angles, $t$ the  Euclidean distance between the two object centers and $p_t$ is a penalty if the objects do not intersect. $S_t$ is calculated in respect to the size of the two objects, because for small objects a small translation can already have a big impact and vice versa for large objects. So the length of the diagonals $d_i$ of both objects are used to calculate two radii $r_i$.

To adjust the score $w_s, w_t$ and $w_r$ can be used. They control how strict the individual scores are. We used $w_s = 0.3$, $w_t = 1$ and $w_r = 0.5$

All the previous scores are in the interval $[0, 1]$ and can be combined into the final score ($S_{srt}$) using a simple weighted average and the penalty $p_t$.
\begin{gather}
S_{srt} =  p_t \cdot (\alpha\;S_s + \beta\;S_t + \gamma\;S_r)\\
\alpha + \beta + \gamma = 1\nonumber
\end{gather}

Using $\alpha, \beta, \gamma$ the weight of the three sub scores can be defined. We used $\gamma = 0.4$ and $\alpha = \beta = 0.3$ to give more weight to the angle, because translation and scaling are easier to learn for the network.

\textit{SRTs} perfectly lines up with the three subtasks (rotation, position, size) a network has to do in order to predict 3D boxes with a yaw angle. It is designed so it can be parametrized to approximate the IoU but considers object orientations.
Using all the parameters the score can be adjusted to suit the needs of the problem.

\subsection{Extended target model in the LMB RFS}
To apply the RFS approach to the output of the YOLO network, which consists of boxes in the three dimensional space, we interpret the output as Gaussian noise corrupted measurements $z_t^i$, $i \in \{1, \dots, N^z_t\}$ of the positional parameters (see Eq. \ref{eq_measurement_set}) and extend those as extended targets \cite{ExtendedObjectTracking} $x_t^i$, $i \in \{1, \dots, N^x_t\}$ (see Eq. \ref{eq:target_set}) with the measurement noise covariance matrix $R = \text{diag}(0.5^2,0.5^2,0.5^2,0.5^2,0.5^2,0.1^2)$.
The target is assumed to move according to a coordinated turn motion model \cite{EKFUKF} with the process noise covariance $Q = \text{diag}(\sigma_a^2,\sigma_\alpha^2)$
consisting of the standard deviation of the acceleration $\sigma_a = 17.89$ and the yaw rates derivative $\sigma_\alpha = 1.49$.

The individual measurements $z$ consist of the box center position $c=[x,y,z]$ in the three dimensional space, the box dimension (length, width, height) $s=[l,w,h]$ and the box orientation along the first dimensions $\phi_t^i$ (yaw), such that 
\begin{equation}
z = \left[c,s,\phi\right].
\end{equation}

The extended target state mean $\bar{x}_t^i$ used for tracking contains the same parameters as the measurements as well as motion parameters of the coordinated turn model consisting of the velocity $v$ and yaw rate $\dot{\phi}$. The state mean of the $i$th target at time $t$  can be described as
\begin{equation}
\bar{x}_t^i = [c_t^i, s_t^i, \phi_t^i, v_t^i,\dot{\phi}_t^i]
\end{equation}
with the according state covariance matrix $\bar{P}_t^i$. We can state the measurement equation
\begin{equation}
z = H\cdot \bar{x}
\end{equation}
for an individual measurement $z$ and the target mean $\bar{x}$, with the measurement matrix
\begin{equation}
H = \begin{pmatrix}
\text{I}_7 & 0\\
\end{pmatrix} \in \mathbb{R}^{7 \times 9},
\end{equation}
where $\text{I}_7$ is the identity matrix of dimension 7.
Based on this measurement equation, a Bayesian filter can be defined where the innovation is calculated using a Kalman filter update according to the stated measurement model. The prediction is performed using an Unscented Kalman filter according to the assumed nonlinear coordinated turn model \cite{EKFUKF}.

In the LMB update step each predicted target is associated with each measurement of the time step and an update according to the defined measurement model is performed. A heatmap is generated from the update likelihood, modeling the association probabilities based on which targets will be kept or discarded. Be $p_a(x^i, z^j)$ the association probability of the measurement $z^j$ and the state $x^i$. If the non-assignment probability   
\begin{align}
p_{na}(z^j) = 1 - \sum_{x^i \in \mathbf{X}} p_a(x^i, z^j)
 \end{align}
is higher than a threshold $P_{na}$, we assume that a new target is born from an unexplained measurement.

The number $N_e$ of targets to be extracted is derived from the mean of the cardinality distribution presented in Eq. \ref{eq_card_dist}. All $N_e$ targets with the highest existence probability $r^{(l)}$ are extracted.  
 
\section{Experiments}
We evaluate Complexer-YOLO on the KITTI benchmarks for 3D object detection and bird's eye view (BEV) detection. Furthermore, we evaluate the capabilities of our multi target tracking with the help of the object tracking benchmark. Our ablation studies investigate the importance of different input features encoded in our voxel representation and show further findings. Finally, some qualitative results visualize the outcome of our model.

\subsection{Training Details}
The KITTI dataset \cite{KITTI} consists of $7481$ training images and $7518$ testing images. First, we follow \cite{MV3D} and use the training/validation split to optimize our settings. Afterwards, we use the full training set for the official evaluation. We augment the training dataset with rotation and increase the size by a factor of $4$. Therefore, we randomly pick $3$ angles between $[-20,20]\deg$ with a minimum difference of $8\deg$ to each other. Similar to \cite{PIXOR}, we use random flipping along the $x$ axis during training.

For training, we use an extended version of the darknet framework \cite{Darknet}. We train the model from scratch for 140k iterations with learning rate scaled at 20k, 80k and 120k iterations respectively.

\subsection{Detection Results}
We submitted our results to the KITTI vision benchmark suite \cite{KITTI} for Orientation Similarity, BEV, 3D Object Detection and Object Tracking benchmarks on the official test set. To achieve a fair comparison, we only selected some of the leading 3D object detectors that are able to detect at least classes \textit{Car}, \textit{Pedestrian} and \textit{Cyclist}. For tracking, only online methods are listed.

We show evaluation results for Orientation Similarity, BEV and 3D object detection in Table \ref{Tab:PerformanceComparison}. Table \ref{tab_tracking_comparison} shows our results in MOT accurancy and precision (MOTA and MOTP), mostly tracked (MT) and mostly lost (ML).

\begin{figure}[!htb]
\centering
\includegraphics[width=0.34\textwidth]{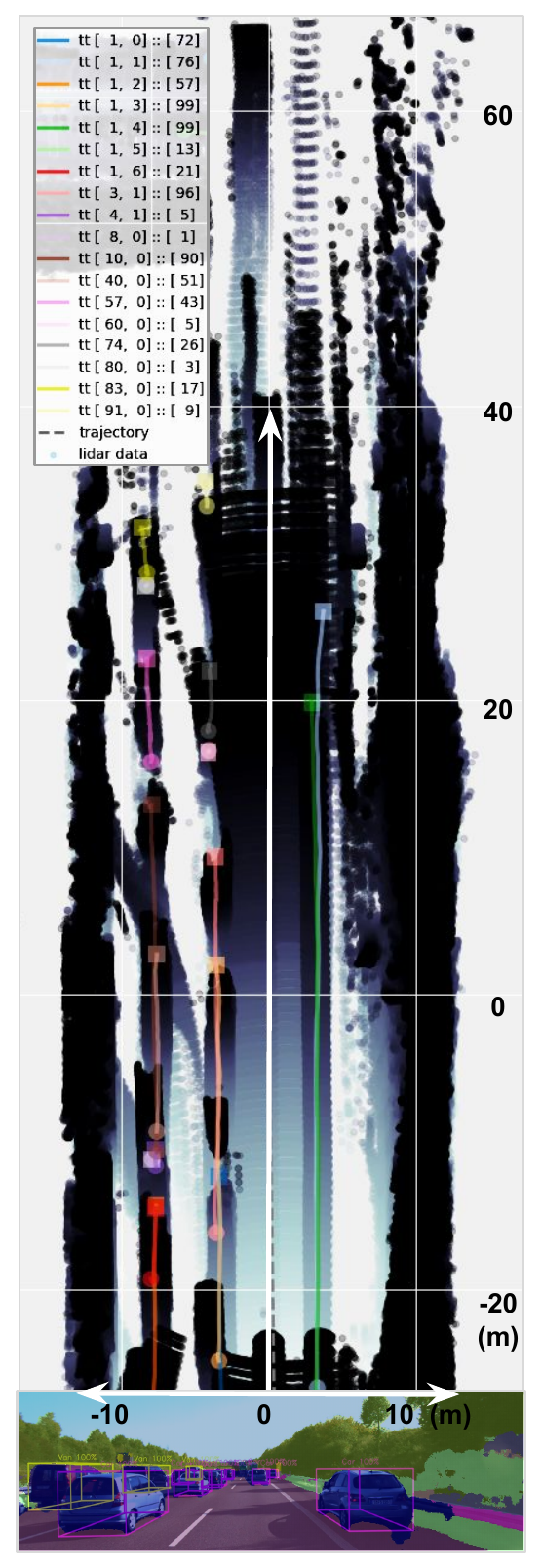}
\caption{Tracked objects trajectories.}
\label{fig:Tracking}
\end{figure}

Unfortunately, the whole evaluation process is based on 2D bounding boxes in camera space due to the handling of \textit{Dontcare} labels and ignored objects, e.g. truncated or occluded (see \cite{KITTI}). Following the 2D Object Detection Benchmark, which is accompanied with BEV and 3D, we achieve $79.31\%$ for class \textit{Car} in \textit{moderate} difficulty. Also, Orientation for these settings is ranked at $79.08\%$. However, our algorithm detects and tracks bounding boxes in 3D space. Therefore, all detections are projected to the image plane. Although we do not track in image space, we achieve state-of-the-art results while running in real-time using our tracking (visualization Fig. \ref{fig:Tracking}). Moreover we are the first one with 3D tracking based on point cloud detections on the KITTI tracking benchmark. But there is an inconsistency compared to BEV and 3D results, where we achieve only $66.07\%$ and $49.44\%$ respectively. Based on less than $50\%$ AP in 3D space, tracking is not able to reach actual results, which we think mainly comes from wrongly counted \textit{Dontcare} objects. In opposition to the KITTI guidelines, we found that their current object detection evaluation scripts fully ignore \textit{Dontcare} labels for BEV and 3D Object detection benchmarks. All such detections count as false positives, which is crucial in our case. Furthermore, most 2D ground truth bounding boxes for class \textit{Pedestrian} are manually refined and do not match a reprojected bounding box from 3D space anymore, which leads to additional wrongly counted false positives, when ignored objects are assigned.

\begin{figure*}[!htb]
\centering
\includegraphics[width=0.80\textwidth]{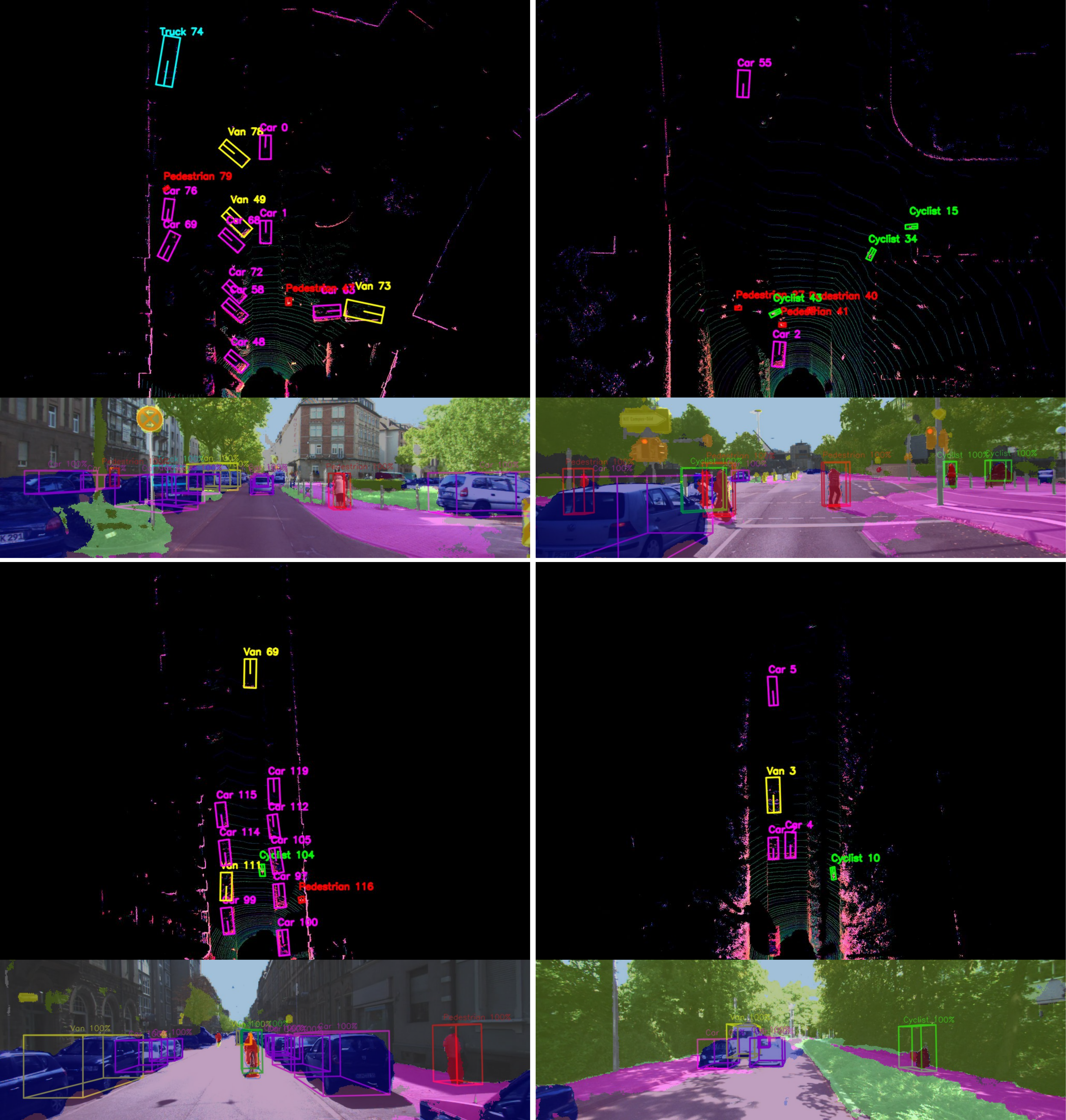}
\caption{Qualitative results. For visualization, our detections are projected into camera space with overlayed pixel wise semantic segmentation.}
\label{fig:QualResults}
\end{figure*}

Fig. \ref{fig:QualResults} shows outstanding results on several sequences with different use cases. Our model is able to detect accurate rotated bounding boxes in 3D space for multiple classes even though the strongly unbalanced dataset. With the help of voxelized semantic features, the network is able to detect even small objects like pedestrians or cyclist, as long as they have a minimum spatial distance to other appearing objects.
\begin{table}[!htb]
\begin{center}
\renewcommand{\arraystretch}{1.1}
\resizebox{\linewidth}{!}{%
\begin{tabular}{|>{\centering\arraybackslash}m{2.9cm}|>{\centering\arraybackslash}m{0.8cm}||>{\centering\arraybackslash}m{0.8cm}|>{\centering\arraybackslash}m{0.8cm}|>{\centering\arraybackslash}m{0.8cm}||>{\centering\arraybackslash}m{0.8cm}|>{\centering\arraybackslash}m{0.8cm}|>{\centering\arraybackslash}m{0.8cm}||>{\centering\arraybackslash}m{0.8cm}|>{\centering\arraybackslash}m{0.8cm}|>{\centering\arraybackslash}m{0.8cm}|}
\hline 
\multirow{2}{*}{\large{\textbf{Method}}} & \multirow{2}{*}{\large{\textbf{FPS}}} & \multicolumn{3}{c||}{\large{\textbf{Car}}} & \multicolumn{3}{c||}{\large{\textbf{Pedestrian}}} & \multicolumn{3}{c|}{\large{\textbf{Cyclist}}} \\ 
 & & \textbf{Easy} & \textbf{Mod.} & \textbf{Hard} & \textbf{Easy} & \textbf{Mod.} & \textbf{Hard} & \textbf{Easy} & \textbf{Mod.} & \textbf{Hard} \\
\hline
\hline
\multicolumn{11}{|c|}{\large{Orientation}} \\
\hline
AVOD-FPN \cite{AVOD} & 10.0 & \textbf{89.95} & \textbf{87.13} & \textbf{79.24} & \textbf{53.36} & \textbf{44.92} & \textbf{43.77} & 67.61 & \textbf{57.53} & 54.16 \\
\hline
SECOND \cite{SECOND} & \textbf{20.0} & 87.84 & 81.31 & 71.95 & 51.56 & 43.51 & 38.78 & \textbf{80.97} & 57.20 & 55.14 \\
\hline
BirdNet \cite{BirdNet} & 9.1 & 50.85 & 35.81 & 34.90 & 21.34 & 17.26 & 16.67 & 41.48 & 30.76 & 28.66 \\
\hline
\hline
Complexer-YOLO & 15.6 & 87.97 & 79.08 & 78.75 & 37.80 & 31.80 & 31.26 & 64.51 & 56.32 & \textbf{56.23} \\
\hline
\hline
\multicolumn{11}{|c|}{\large{BEV}} \\
\hline
F-PointNet \cite{FrustumPointNets} & 5.9 & 88.70 & \textbf{84.00} & 75.33 & 58.09 & 50.22 & 47.20 & \textbf{75.38} & \textbf{61.96} & \textbf{54.68} \\
\hline
AVOD-FPN \cite{AVOD} & 10.0 & 88.53 & 83.79 & \textbf{77.90} & \textbf{58.75} & \textbf{51.05} & \textbf{47.54} & 68.09 & 57.48 & 50.77 \\
\hline
VoxelNet \cite{VoxelNet} & 4.4 & \textbf{89.35} & 79.26 & 77.39 & 46.13 & 40.74 & 38.11 & 66.70 & 54.76 & 50.55 \\
\hline
BirdNet \cite{BirdNet} & 9.1 & 75.52 & 50.81 & 50.00 & 26.07 & 21.35 & 19.96 & 38.93 & 27.18 & 25.51 \\
\hline
\hline
Complexer-YOLO & \textbf{15.6} & 74.23 & 66.07 & 65.70 & 22.00 & 20.88 & 20.81 & 36.12 & 30.16 & 26.01 \\
\hline
\hline
\multicolumn{11}{|c|}{\large{3D}} \\
\hline
\hline
F-PointNet \cite{FrustumPointNets} & 5.9 & 81.20 & 70.39 & 62.19 & \textbf{51.21} & \textbf{44.89} & 40.23 & \textbf{71.96} & \textbf{56.77} & \textbf{50.39} \\
\hline
AVOD-FPN \cite{AVOD} & 10.0 & \textbf{81.94} & \textbf{71.88} & \textbf{66.38} & 50.80 & 42.81 & \textbf{40.88} & 64.00 & 52.18 & 46.61 \\
\hline
VoxelNet \cite{VoxelNet} & 4.4 & 77.47 & 65.11 & 57.73 & 39.48 & 33.69 & 31.51 & 61.22 & 48.36 & 44.37 \\
\hline
BirdNet \cite{BirdNet} & 9.1 & 14.75 & 13.44 & 12.04 & 14.31 & 11.80 & 10.55 & 18.35 & 12.43 & 11.88 \\
\hline
\hline
Complexer-YOLO & \textbf{15.6} & 55.63 & 49.44 & 44.13 & 19.45 & 15.32 & 14.80 & 28.36 & 23.48 & 22.85 \\
\hline
\end{tabular}}
\end{center}
\caption{Evaluation of orientation, bird's eye view and 3D detection. Frames per second (FPS) and APs (in \%) on KITTI test set.}
\label{Tab:PerformanceComparison}
\end{table}

\begin{table}[!htb]
\begin{center}
\renewcommand{\arraystretch}{1.1}
\resizebox{\linewidth}{!}{%
\begin{tabular}{|l|c|c|c|c|c|}
\hline
Method & MOTA [\%] & MOTP [\%] & MT [\%] & ML [\%] & FPS\\
\hline\hline
MOTBeyondPixels \cite{BeyondPixelsMultiObjectTracking} & \textbf{84.24} & \textbf{85.73} & \textbf{73.23} & \textbf{2.77} & 3.3 \\
IMMDP \cite{BeyondPixelsMultiObjectTracking} & 83.04 & 82.74 & 60.62 & 11.38 & 5.3 \\
3D-CNN/PMBM \cite{MultiObjectTrackingDeepLearningPMBMFilterung} & 80.39 & 81.26 & 62.77 & 6.15 & \textbf{100} \\
mbodSSP \cite{FollowMe} & 72.69 & 78.75 & 48.77 & 8.77 & \textbf{100} \\
\hline
\hline
Ours & 75.70 & 78.46 & 58.00 & 5.08 & \textbf{100}\\
\hline
\end{tabular}}
\end{center}
\caption{Comparison with non-anonymous pure online submissions on KITTI MOT benchmark \cite{KITTI}.}
\label{tab_tracking_comparison}
\end{table}

\subsection{Ablation Study}
We conducted ablation experiments with fixed training setup to investigate the influence of our hyper parameters and several input features. The use of $21$ height channels for our voxel map results in similar mAP at IoU threshold $0.7$ as using $51$ channels (cuboidal voxels). It seems that our network is not able to fully utilize fine grained height information. Furthermore, it is the best trade-off for runtime and accuracy, because runtime was slightly increasing for more than $21$ height channels with our hardware setup. Table \ref{tab:ablation} shows results using different voxel maps with intensity values from Lidar sensor normalized in range $[1, 2]$, binary occupancy similar to \cite{FastAndFurious} and our novel semantic map. Additionally, the approach from \cite{ComplexYolo} using extracted features encoded as an RGB image is listed.

\begin{figure}[!tb]
\centering
\includegraphics[width=0.43\textwidth]{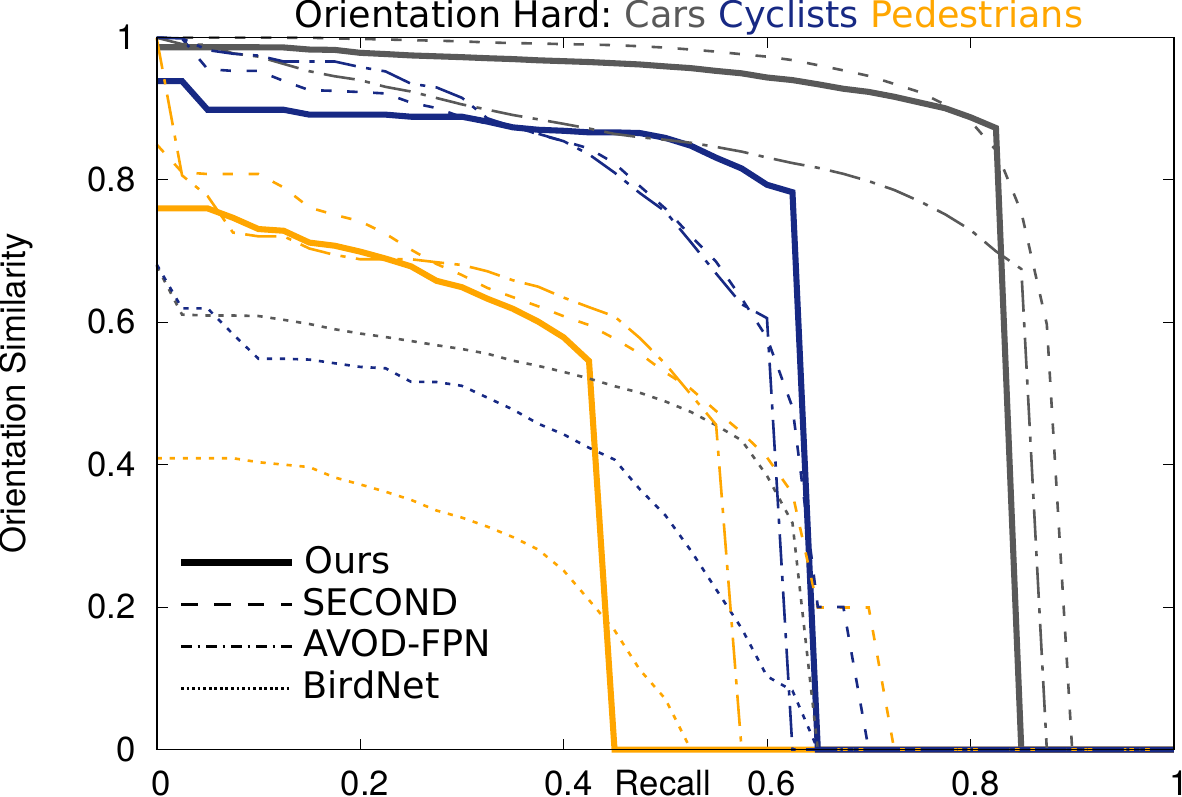}
\caption{Results for the orientation benchmark compared to SECOND \cite{SECOND}, BirdNet \cite{BirdNet}, AVOD-FPN \cite{AVOD} on official KITTI test set.}
\label{fig:Orientation}
\end{figure}

\begin{figure}[!tb]
\centering
\includegraphics[width=0.44\textwidth]{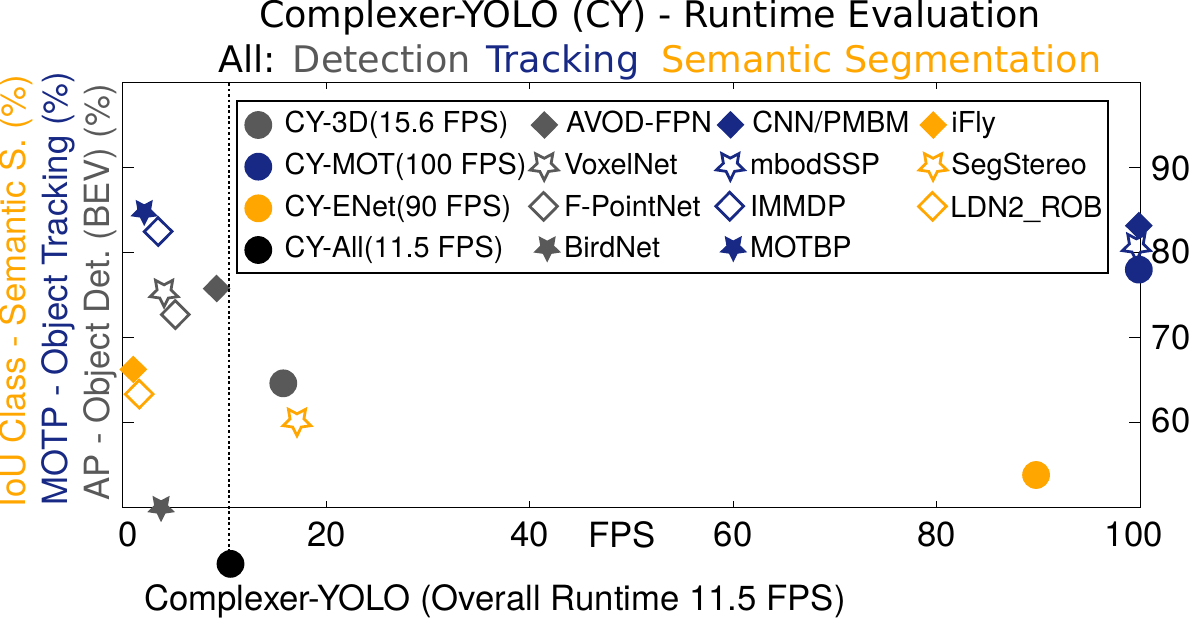}
\caption{The Complexer-YOLO runtime evaluation (reference hardware: NVIDIA GTX1080i/Titan) shows state-of-the-art results of all single tasks (Semantic Segmentation, 3D object detection (BEV, cars $\rightarrow$ hard, Tab.~\ref{Tab:PerformanceComparison}), Tracking Tab.~\ref{tab_tracking_comparison}). Results for Semantic Segmentation are taken from the KITTI leader board. We point out, that our overall Detection and Tracking Pipeline is faster than many single task algorithms.}
\label{fig:Runtime}
\end{figure}

In order to reduce wrongly counted false positives due to ignored \textit{Dontcare} labels, we tried to filter our detections in a post processing step. Therefore, we counted the number of 3D points falling into each 3D bounding box. All detections with less then $13$ points and less than $52m$ radial distance to the Lidar sensor were removed, because \textit{Dontcare} is often used for objects at higher distances or occluded objects. This improved all object detection results by $1.3\%$ on average, but decreased e.g. BEV for \textit{Car} on \textit{moderate} difficulty by $4.8\%$. Consequently, our filter removed a few \textit{Dontcare} or ignored detections, but removed correct ones as well. Also, it seemed to have stronger impact on \textit{moderate} settings since valid \textit{easy} detections are all in near range, which explains AP drops from \textit{easy}.

Finally, using \textit{SRTs} for training instead of IoU gives $1.3\%$ improvement on mAP at IoU $0.7$ as it directly penalizes orientation. It also halved our training time and resulted in a $10$-$20\%$ runtime improvement for inference.

Additionally, we tried to limit object rotations into subsections using anchors instead of complex angles for the full $360\deg$, but this decreased accuracy. Further investigation is required here, because we see a potential reduction in complexity for the learning task of the network.

\begin{table}[!htb]
\begin{center}
\resizebox{0.45\linewidth}{!}{%
\begin{tabular}{|l|c|c|}
\hline
Feature & IoU 0.7 & SRTs 0.7 \\
\hline\hline
RGB & 28.64 & 30.02 \\
Occupancy & 31.93 & 33.24 \\
Intensity & 32.39 & 33.57\\
Semantic & 34.14 & 35.43 \\
\hline
\end{tabular}}
\end{center}
\caption{Ablation study of different input features. mAP values (in \%) for the 3D benchmark on KITTI validation set.}
\label{tab:ablation}
\end{table}

\section{Conclusion}
In this work we propose Complexer-YOLO, a tracked real-time 3D object detector that operates on point clouds fused with visual semantic segmentation. Our architecture takes advantage of both spatial Lidar data and explored scene understanding from 2D. Detection results obtained from 3D space show competitive performance on KITTI benchmarks \cite{KITTI} compared to state-of-the-art. At the same time, we introduce \textit{SRTs}, a powerful, more flexible and simplified evaluation metric for object comparison that overcomes the limits of IoU. 

\clearpage

{\small
\bibliographystyle{ieee}
\bibliography{egbib}
}

\end{document}